\def\year{2021}\relax
\documentclass[letterpaper]{article} 
\usepackage{aaai21}  
\usepackage{times}  
\usepackage{helvet} 
\usepackage{courier}  
\usepackage[hyphens]{url}  
\usepackage{graphicx} 
\usepackage{natbib}  
\usepackage{caption} 
\usepackage{comment}
\usepackage{ifthen}
\newcommand{\boldhdr}[1]{\noindent \textbf{#1.}}

\newcommand{\showcomments}{yes}

\newcommand\animesh[1]{
    \ifthenelse{\equal{\showcomments}{yes}}{{\color{red} [Animesh: #1]}}{\ignorespaces}
}

\newcommand\ming[1]{
    \ifthenelse{\equal{\showcomments}{yes}}{{\color{magenta} [Ming: #1]}}{\ignorespaces}
}

\newcommand\kaiqi[1]{
    \ifthenelse{\equal{\showcomments}{yes}}{{\color{blue} [Kaiqi: #1]}}{\ignorespaces}
}

\usepackage{amsmath}
\usepackage{color} 
\usepackage{mathabx}
\usepackage{booktabs}
\usepackage{multirow}
\usepackage{stfloats}
\usepackage{subfigure}
\usepackage{float}
\usepackage{lipsum} 
\usepackage{fancyhdr}
\usepackage{cite}
\usepackage{algorithmic}
\usepackage{graphicx}
\usepackage{textcomp}
\usepackage{xcolor}
\usepackage{multirow}
\usepackage{multicol}

\title{Iterative Activation-based Structured Pruning}
\author{
    Kaiqi Zhao\thanks{This work is conduct during Kaiqi's internship at Amazon.com Inc.}, Arizona State University (AWS Intern 2020)\\
    Animesh Jain, AWS \\
    Ming Zhao, Arizona State University 
    
}
\affiliations{
}

\begin{document}
\pagestyle{plain}
\maketitle



\begin{abstract}
Deploying complex deep learning models on edge devices is challenging because they have substantial compute and memory resource requirements, whereas edge devices' resource budget is limited. To solve this problem, extensive pruning techniques have been proposed for compressing networks. Recent advances based on the Lottery Ticket Hypothesis (LTH) show that iterative model pruning tends to produce smaller and more accurate models. However, LTH research focuses on unstructured pruning, which is hardware-inefficient and difficult to accelerate on hardware platforms.

In this paper, we investigate iterative pruning in the context of structured pruning because structurally pruned models map well on commodity hardware. We find that directly applying a structured weight-based pruning technique iteratively, called iterative L1-norm based pruning (ILP), does not produce accurate pruned models. To solve this problem, we propose two activation-based pruning methods, Iterative Activation-based Pruning (IAP) and Adaptive Iterative Activation-based Pruning (AIAP). We observe that, with only 1\% accuracy loss, IAP and AIAP achieve 7.75$\times$ and 15.88$\times$ compression on LeNet-5, and 1.25$\times$ and 1.71$\times$ compression on ResNet-50, whereas ILP achieves 4.77$\times$ and 1.13$\times$, respectively.

\end{abstract}
\section{Introduction}\label{introduction}
Deep neural networks (DNNs) have substantial compute and memory requirements. As deep learning becomes pervasive and moves towards edge devices, DNN deployment becomes harder because of the mistmatch between resource-hungry DNNs and resource-constrained edge devices~\cite{li2018edge, li2019edge}. Deep learning researchers and practitioners have proposed many techniques to alleviate this resource pressure~\cite{chu2003structured, han2015deep, polino2018model, yim2017gift, pham2018efficient}. Among these efforts, DNN pruning is a promising approach~\cite{li2016pruning, han2015deep, molchanov2016pruning, theis2018faster, renda2020comparing}, which identifies the parameters (or weight elements) that do not contribute significantly to the accuracy, and prunes them from the network.

\subsection{Unstructured vs. Structured Pruning}
Like other compression techniques, pruning has a performance-accuracy tradeoff, i.e, as we compress the model more aggressively, we achieve better hardware performance (higher reductions in computation and memory requirements) but suffer higher accuracy losses. In this paper, we  measure  performance across two dimensions. First,  \emph{memory footprint}, i.e., memory space used to store parameters, and second, \emph{inference latency}, i.e., time to run one inference. Pruning can be broadly categorized into two classes, namely, unstructured pruning and structured pruning. The two techniques take different approaches on memory footprint reduction, and impact inference latency and accuracy differently.

Unstructured pruning is a fine-grained approach that prunes individual unimportant elements in weight tensors. It allows better accuracy with higher memory footprint reduction, compared to structured pruning, as it selectively prunes unimportant elements from a weight filter while retaining the important elements. However, unstructured pruning is hardware-inefficient because it is difficult to map random/indirect memory in the pruned weight representation on general-purpose hardware platforms like Intel/ARM CPU vector units or NVDIA GPU CUDA cores~\cite{he2018amc}, e.g,~\cite{deftnn} shows directly applying NVDIA cuSPARSE on unstructured pruned models can lead to 60$\times$ slowdown on GPU compared to dense kernels.

On the other hand, structured pruning is a coarse-grained approach that prunes entire regular regions of weight tensors (e.g., filters or output channels)~\cite{han2015deep}. It tends to cause higher drop in accuracy than unstructured pruning because by removing entire regions, it might remove weight elements that are important to the final accuracy~\cite{li2016pruning}. However, structurally pruned models can be mapped easily to general-purpose hardware and accelerated directly with off-the-shelf hardware and libraries~\cite{he2018amc}.

\subsection{Iterative Pruning}
Another choice in DNN pruning is between one-shot pruning and iterative pruning. One-shot pruning prunes a pre-trained model and then retrains it \emph{once} to make up for the accuracy loss. Iterative pruning, on the other hand, prunes and retrains the model in multiple rounds. Both techniques can choose either structured or unstructured pruning techniques. Recent research, particularly the Lottery Ticket Hypothesis (LTH)~\cite{frankle2018lottery}, shows that iterative pruning leads to smaller and more accurate models, when weights or/and learning rate are rewound to an early training iteration after each pruning round (discussed more in Background and Related Works\ref{sec:related_work}). However, majority of LTH and the follow-up research has been done with unstructured pruning which is hardware-inefficient. In this paper, we investigate hardware-friendly \textbf{iterative structured pruning with rewinding}.

Prior iterative unstructured pruning with rewinding research (LTH and follow-up papers) uses widely adopted iterative magnitude pruning (IMP) technique~\cite{han2015learning}. To understand the state of iterative structured pruning, we experiment with IMP's structured pruning counterpart---L1-norm structured pruning (ILP)~\cite{li2016pruning}. ILP removes entire filters depending on their L1-norm value. However, we observe that ILP leads to severe accuracy losses with small reduction in memory size, e.g., ILP could compress ResNet-50 by only 1.13$\times$ when the maximum accuracy loss is limited to 1\% (higher accuracy losses tend to be unsuitable for industrial deployments). Therefore, directly applying iterative pruning to existing structured pruning methods does not produce accurate pruned models.

\subsection{Our Solution}
Our results from ILP show that just using weights values to identify and prune unimportant filters is not effective. Sometimes, filters that have small weight values can also produce useful non-zero activation patterns that are important for learning features during backpropagation. Instead, we propose to use activation values, i.e., the intermediate output tensors after the non-linear activation, to identify unimportant filters. This tends to be more effective because activation values not only capture features of training dataset, but also contain the information of convolution layers that act as feature detectors for prediction tasks.

Based on this insight, we propose two activation-based structured pruning methods---Iterative Activation-based pruning (IAP) and Adaptive Iterative Activation-based pruning (AIAP). IAP prunes a fixed percentage of unimportant filters every round depending on the mean of the activation values. AIAP prunes filters if the mean activation value is below an adaptive threshold that increases as the model gets smaller every round. 

In this paper, we seek to answer following questions in the context of iterative structured pruning with rewinding:

\begin{itemize}
    \item \emph{Q1:} Can iterative structured pruning with rewinding find lottery tickets, i.e., generate pruned models that are smaller but also at least as accurate as the original model?\\ \emph{Answer:} Yes, we observe winning lottery tickets for both shallow and deep networks. IAP and AIAP compress LeNet-5 by 2.55$\times$ and 3.95$\times$ and ResNet-50 by 1.11$\times$ and 1.17$\times$, respectively, with no accuracy loss. 
    \item \emph{Q2:} What is the accuracy and memory footprint tradeoff for iterative structured pruning?\\
    \emph{Answer:} We observe that IAP and AIAP compress the models with graceful accuracy degradation, e.g., for an acceptable accuracy loss of 1\%, IAP and AIAP achieve 1.25$\times$ and 1.71$\times$ compression, respectively, compared to just 1.13$\times$ for ILP.
    \item \emph{Q3:} What is the accuracy and inference latency tradeoff for iterative structured pruning?\\
    \emph{Answer:} We observe that IAP and AIAP reduce the compute flops by 1.5$\times$ and 1.6$\times$ for LeNet-300-100 and 1.05$\times$ and 1.33$\times$ for ResNet-50 compared to ILP, when accuracy loss is 1\%. As our pruning techniques are regular and remove filters, we can employ off-the-shelf libraries to speedup the model inference.\vspace{-3pt}
\end{itemize}

To the best of our knowledge, the paper is the first to investigate iterative structured pruning with weight and learning rate rewinding. We introduce two activation-based structured pruning techniques to prune the models. The models pruned by the proposed techniques produce a new baseline for the future work on structurally pruned models.

\section{Background and Related Works}
\label{sec:related_work}

Deep learning researchers and practitioners have proposed many techniques to reduce the resource-hungry requirements of deep neural networks, such as low-rank approximation~\cite{chu2003structured, clarkson2017low}, quantization~\cite{lan2014matrix, polino2018model}, weight sharing~\cite{han2015deep, kadetotad2016efficient}, knowledge distillation~\cite{polino2018model, yim2017gift}, neural architecture search~\cite{zoph2016neural, pham2018efficient} and pruning~\cite{li2016pruning, han2015deep, srinivas2015data, molchanov2016pruning}. There has also been substantial work in manually designing new model topology, like MobileNet~\cite{howard2017mobilenets} and EfficientNet~\cite{tan2019efficientnet}, that are suitable for edge device deployment but are less accurate compared to traditional models like ResNet~\cite{he2016deep}. Model pruning is gaining popularity because researchers are gradually building an \emph{automatic} flow of pruning the models, reducing the need of manual rules or significant domain expertise. 

\smallskip
\boldhdr{Unstructured Pruning} One of the earliest works in unstructured pruning is Optimal Brain Damage, which prunes weights based on Hessian of the loss function~\cite{lecun1990optimal}. Recently, Han proposed to prune weights with smallest magnitude values~\cite{li2016pruning}, which has been widely adopted.

Research has shown that traditional sparse computation libraries, like cuSPARSE or MKLDNN, observe substantial slowdown on unstructured pruned models because the libraries are designed for extremely high sparsity ($>$99\%), while unstructured sparsity leads to much lower 75\%-90\% sparsity~\cite{deftnn}. To solve this, there are initial efforts to manually write kernels suitable for this sparsity~\cite{sparse_conv}. But overall, unstructured pruned models are still hard to accelerate on commodity hardware. 

\smallskip
\boldhdr{Structured Pruning} Earlier pruning research focused on manually finding heuristics on how to prune, such as L1-norm~\cite{li2016pruning}, average percentage of zero~\cite{molchanov2016pruning}, and other information considering the relationship between neighboring layers~\cite{theis2018faster, lee2018snip}. Recently, the focus has shifted to finding the marginal filters \emph{automotically}. Network slimming prunes filters with lower scaling factors, producing automatically discovered target architectures~\cite{yu2018slimmable}. AutoML uses reinforcement learning to iteratively prune the convolution layers~\cite{he2018amc}. Though time consuming, automatic pruning has produced smaller models than manual techniques for similar accuracy. 

Structured pruned models are easier to accelerate as they remove entire regions of the weight tensor. For example, L1-norm filter pruning prunes filters, thus reducing the number of elements in the filter dimension of weight tensors. Essentially, this results in convolving two smaller tensors, which can be accelerated easily using off-the-shelf libraries. This is much harder to do in unstructured pruning.

\vspace{2pt}
\boldhdr{Iterative Pruning with Rewinding} Lottery Ticket Hypothesis (LTH) is a recent work on iterative pruning, which introduces rewinding with iterative pruning to generate smaller and accurate models~\cite{frankle2018lottery}. LTH posits that a dense randomly initialized network has a subnetwork, termed as a winning ticket, which can be trained in isolation to achieve better or at least similar accuracy than the original network~\cite{frankle2018lottery, frankle2019stabilizing}. While pruning the model iteratively, it rewinds the weights and learning rate to some early epoch in the training phase.

There are two types of rewinding---weight rewinding and learning rate rewinding~\cite{renda2020comparing}. In weight rewinding, both weights and learning rate are rewound to some early training epoch. In learning rate rewinding, only the learning rate is rewound. However, rewinding has only been validated in the context of unstructured pruning, which as mentioned above is hardware-inefficient. 

In this paper, we focus on iterative structured pruning with rewinding. We aim to generate pruned models that are not only smaller and accurate, but also faster.

\section{Methodology}\label{methodology}

We first present an overview of iterative pruning flow, and then detail the design of activation-based structured pruning.

\begin{figure}[t]
	\centering
	\includegraphics[width=6.5cm]{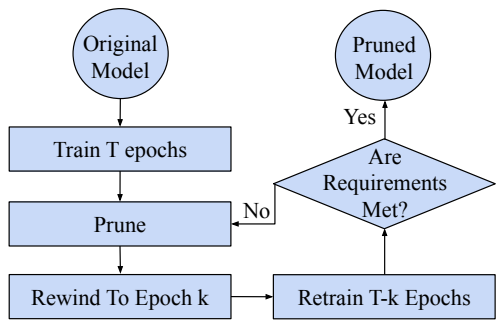}\vspace{-6pt}
	\caption{Iterative pruning flow with rewinding}
	\label{fig:pruning_flow}
 	\vspace{-9pt}
\end{figure}

\begin{table*}[t]
\small
\centering
\begin{tabular}{@{}l@{}}
\toprule
\textbf {Algorithm 1 Iterative Structured Pruning Algorithm}  \\ \midrule
\hspace{2mm} 1: {[}Initialize{]} Randomly initialize a neural network $f(x; M^0 \odot W_{0}^{0})$ with initial pruning mask $M^0=\left \{ 0,1 \right \}^{\left | W_{0}^{0} \right |}$. \\
\hspace{2mm} 2: {[}Save weights at epoch k{]} Train the network for k epochs, producing network $f(x; M^0 \odot W_{k}^{0})$, and save weights $W_{k}^{0}$.\\
\hspace{2mm} 3: {[}Train to Completion{]} Train the network for T-k further epochs to converge,  producing network $f(x; M^0 \odot W_{T}^{0})$. \\
\hspace{2mm} For each pruning round r (r\textgreater{}=1): \\
\hspace{6mm} 4: {[}Prune{]} Prune filters from $W_{T}^{r-1}$, producing a new mask $M^r$, and a network $f(x; M^r \odot W_{T}^{r-1})$. \\
\hspace{6mm} 5: {[}Rewind Weights{]} Reset the remaining filters to their values in $W_{k}^{0}$ at epoch $k$, producing network $f(x; M^r \odot W_{k}^{r-1})$. \\
\hspace{6mm} 6: {[}Rewind Learning Rate{]} Reset the learning rate schedule to its state from epoch k. \\
\hspace{6mm} 7: {[}Retrain{]} Retrain the unpruned filters for T-k epoch to converge, producing a network $f(x; M^r \odot W_{T}^{r})$.\\ 
\bottomrule
\end{tabular}
\end{table*}

\subsection{Flow of Iterative Pruning}
We present the high-level flow of iterative pruning in Figure~\ref{fig:pruning_flow}. First, train the model, say for $T$ epochs to create the original model. Next step is to figure out what to prune. Either structured or unstructured pruning technique can be used to identify unimportant weight elements/regions and remove them from the network. By this time, we have a pruned model but its accuracy is likely to be bad.

Next, rewind weights and learning rate or just learning rate to their original values at epoch $k$~\cite{frankle2018lottery}. Then retrain the rewound pruned model for $T-k$ epochs to make up for the accuracy losses introduced by pruning. Check if the retrained pruned model has met our requirement (desired accuracy or memory footprint). If not, repeat the \emph{pruning round}---prune, rewind and retrain---until the requirements are met. 

We present a more formal and complete algorithm in Algorithm 1. To represent pruning of weights, we use a mask $M^r \epsilon \left \{0, 1\right \} ^ d$ for each weight tensor $W_t^r \epsilon R^d$, where $r$ is pruning round number and $t$ is the training epoch. Therefore, the pruned network at the end of training epoch $T$ is represented by the element-wise product $M^r \odot W_T^r$. The first three steps are to train the original model to completion, while saving the weights at epoch $k$. Steps 4--7 represent a pruning round. Step 4 prunes the model. Step 5 (optional) and 6 perform rewinding. Step 7 retrains the pruned model for the remaining $T-k$ epochs.

\begin{figure}[t]
	\centering
	\includegraphics[width=0.45\textwidth]{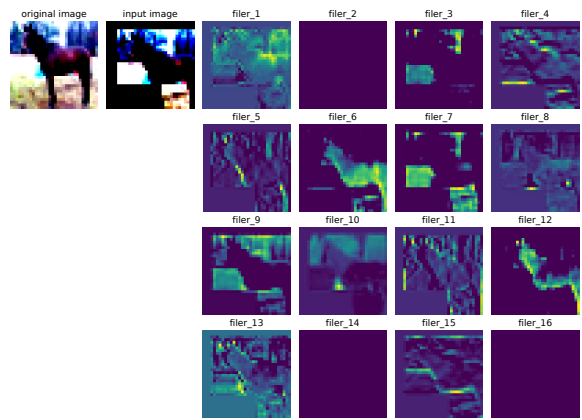}\vspace{-6pt}
	\caption{Activation output of 16 filters of a conv2d layer. On the left, the first image is the original and the second is after data augmentation; right portion shows filter activation outputs.}
	\label{visualization_of_activations}
	\vspace{-0.15in}
\end{figure}

\subsection{Iterative Structured Pruning---Motivation}
In this subsection, we present the motivation behind the design of our pruning techniques. We focus on weight- and activation-based structural pruning techniques to prune filters. Such pruning techniques are simple to implement and work well with iterative pruning, e.g., LTH uses simple unstructured iterative weight-based magnitude pruning (IMP) and successfully finds lottery tickets.

To evaluate the state-of-the-art for iterative structured pruning, we choose simple and widely used L1-norm based structured pruning method. We refer to it as Iterative L1-norm Pruning (ILP). ILP is a weight-based pruning approach that removes filters with lowest L1-norm values. However, we observe that ILP does not produce accurate pruned models. For example, the largest achievable compression ratio of ResNet-50 pruned by ILP is only 1.13$\times$ with 1\% top-1 accuracy loss. Some filters, even though their L1-norm is small, can produce useful non-zero activation values that are important for learning features during backpropagation.

Instead, we observe that activation values are more effective in finding unimportant filters. We present the visual motivation in Figure~\ref{visualization_of_activations}. The figure shows the activation output of 16 filters of a convolution layer on one input image. The first image on the left is the original image, and the second image is the input features after data augmentation. We observe that some filters extract image features with high activation patterns, e.g., $6$th and $12$th filters. In comparison, the activation outputs of some filters are close to zero, such as the $2$nd, $14$th, and $16$th. Therefore, from visual inspection, removing filters with weak activation patterns is likely to have minimal impact on the final accuracy of the pruned model.


Overall, we find that activation-based pruning leads to more accurate pruned models. Activations induce the sparsity, and enable convolutional layers acting as feature detectors. If an activation value is small, then its corresponding feature detector is not important for prediction tasks. So activation values, i.e., the intermediate output tensors after the non-linear activation, not only detect features of training dataset, but also contain the information of convolution layers that act as feature detectors for prediction tasks.


\subsection{Activation-based Structured Pruning}
To solve the aforementioned problem, we introduce two simple activation-based structured pruning methods---Iterative Activation-based Pruning (IAP) and Adaptive Iterative Activation-based Pruning (AIAP). Note that these are structured pruning techniques, and therefore hardware-efficient and easy to accelerate with off-the-shelf libraries.

\begin{figure}[t]
	\centering
	\includegraphics[width=0.45\textwidth]{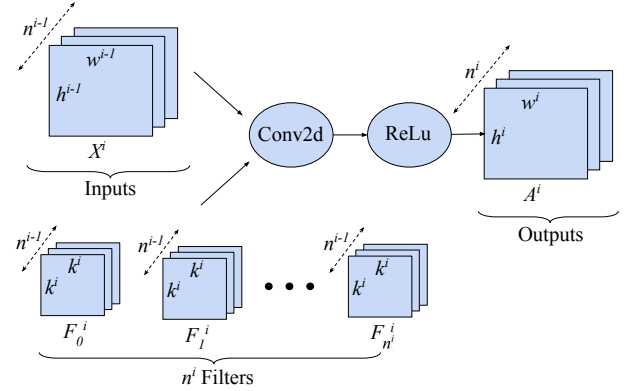}
	\caption{Input and output tensors of a convolution layer. Convolution layer is typically followed by an activation layer like ReLu.}
	\label{fig:conv2d}
 	\vspace{-9pt}
\end{figure}

\subsubsection{Iterative Activation-based Filter Pruning (IAP).}
Figure~\ref{fig:conv2d} shows the inputs and outputs of a 2D convolution layer (referred to as conv2d), followed by an activation layer. For the $i$th conv2d layer, let $X^i \epsilon R^{b \times n^{i-1}\times h^{i-1}\times w^{i-1}}$ denote one batch of the input features, and $F^i_j  \epsilon R^{n^{i-1}\times k^i \times k^i}$ be the $j$th filter, where $b$ denotes batch size, $h_{i-1}/w_{i-1}$ is the height/width of the input features, $n^{i-1}$ is the number of input channels, and $n^{i}$ is the number of output channels. The output of the filter $F^i_j$ after ReLu mapping is therefore denoted by $A^i_j \epsilon R^{b \times h^i \times w^i}$, where $h_{i}/w_{i}$ is the height/width of the output features.  

For pruning, we introduce a mask $M^i_j  \epsilon \left \{ 0, 1 \right \}^{n^{i-1}\times k^i \times k^i}$ for each filter $F^i_j$. The effective weights of the conv2d layer after pruning is the elementwise product of the filter and mask---$F^i_j \odot M^i_j$. For a filter, all values of this mask is either 0 or 1, thus either keeping or pruning the entire filter.

Our structured pruning technique is applied iteratively with rewinding. For each pruning round $r$, we prune $p\%$ filters from each conv2d layer. We present the steps in each pruning round for the $i$th conv2d layer below:

\begin{enumerate}
    \item Select unpruned filters with a non-zero mask: $F^i = \left \{ F^i_j | M^i_j \neq 0 \right \}$, and let $m^i$ denote the number of unpruned filters in $F^i$: $m^i =\left |  F^i \right |$.
    \item For each filter $F^i_j$ in $F^i$, calculate the mean of its activation $A^i_j$: $\bar{A^i_j} = \frac{1}{b}\sum_{k=1}^{h^i}\sum_{l=1}^{w^i}a_{k,l} $, where $a_{k,l}$ denotes every value of $A^i_j$.
    \item Sort filters by $\bar{A^i_j}$.
    \item Prune $p\% \times m^i$ filters with the smallest activation mean values, i.e., let $M^i_j = 0$ if $F^i_j$ is pruned.
\end{enumerate}

\subsubsection{Adaptive Iterative Activation-based Pruning (AIAP).}
Weight-based pruning techniques like ILP and the above proposed IAP prune a fixed percentage of filters in each pruning round. This approach can be too aggressive in some pruning rounds and remove useful filters, making it hard to recover accuracy loss.

In order to avoid removing useful filters, we propose to prune different numbers of filters in each pruning round based on the reduced number of parameters. We introduce an Adaptive Iterative Activation-based pruning (AIAP) method which adapts the aggressiveness of pruning automatically. In contrast to IAP that removes a fixed percentage (p\%) of filters with lowest mean activation values, AIAP removes the filters that are below the adaptive threshold (T). The pruning starts conservatively, with T initialized to 0, and gradually increases its aggressiveness: T increases only if the amount of parameters removed in the previous round is insufficient. We show the procedure for AIAP in a pruning round $r$ for the $i$th conv2d layer as follows:

\begin{enumerate}
    \item Calculate the fraction of parameters of conv2d layers pruned in the last round, denoted as $d$: $d = \frac{P[r-2] - P[r-1]}{P[0]}$, where $P[r-1]$ and $P[r-2]$ are the remaining number of parameters of conv2d layers in pruning round $r-1$ and $r-2$, respectively, and $P[0]$ denotes the total number of parameters of conv2d layers of the original network.
    \item Calculate threshold, denoted as $T[r]$, which is set to 0.0 initially, and then increases gradually ($\lambda$ denotes the increment) if the remaining number of filters does not decrease continuously with pruning:\\ 
    $T[r] = \left\{\begin{matrix}
    0.0           & if \quad r \leq 3 & \\ 
    T[r-1] + \lambda  & if \quad r>3  &and \quad d<1\%\\ 
    T[r-1]         & if \quad r>3  &and \quad d \geqslant 1\%\\
    \end{matrix}\right.$
    \item Select unpruned filters that have a non-zero mask: $F^i = \left \{ F^i_j | M^i_j \neq 0 \right \}$.
    \item For each filter $F^i_j$ in $F^i$, calculate the mean of its activation: $\bar{A^i_j} = \frac{1}{b}\sum_{k=1}^{h^i}\sum_{l=1}^{w^i}a_{k,l} $, where $a_{k,l}$ denotes every value of $A^i_j$.
    \item Prune the filter if its activation mean value is not larger than threshold $T[r]$, i.e., set $M^i_j=0$ if $\bar{A^i_j} \leq T[r]$; Otherwise, set $M^i_j=1$.
\end{enumerate}

\smallskip
Note that both IAP and AIAP are applied iteratively, and stop when the accuracy or memory footprint requirement is met. As shown in Figure~\ref{fig:pruning_flow}, IAP and AIAP are the implementation of the ``Prune'' box. Either weight rewinding and/or learning rate rewinding can be used to implement the ``Rewind" box in the flow.

\section{Evaluation}\label{evaluation}

\subsection{Experimental Setup}
\boldhdr{Deep Neural Networks and Datasets} We evaluate our proposed techniques on a variety of image classification DNNs: LeNet-300-100 on MNIST~\cite{deng2012mnist}, LeNet-5 on CIFAR-10~\cite{krizhevsky2009learning}, and ResNet-50 on ImageNet~\cite{deng2009imagenet}. We present the relevant details in Table~\ref{tab:dnns}. We implement all the iterative techniques on PyTorch version 1.6.0.

\begin{table*}[t]
    \small
	\centering
	\caption{Networks, datasets, and hyperparameters.}\vspace{-6pt}
	\label{tab:hyperparameters}
	\begin{tabular}{lll|ll|l}
		\toprule
		\vspace{0.5mm}
		Dataset & Network & \#Params & Optimizer & \parbox{2.0cm}{Learning Rate \\(t = training epoch)}& \parbox{2.0cm}{Test Accuracy \\(Top-1)}\\
		\hline
 		\vspace{0.5mm}
		MNIST & LeNet-300-100 & 0.27M & \parbox{3.0cm}{Nesterov Adam \\Weight decay: 0.0001\\batch size: 60} 
		& $0.0012 \quad t\epsilon [0, 6)$ & $97.34\% \pm 0.20\%$\\
		
		\hline
        \vspace{0.5mm}
		CIFAR-10 & LeNet-5 & 4.30M & \parbox{3.0cm}{Nesterov Adam \\Weight decay: 0.0001\\batch size: 60} 
		& $0.0002 \quad t\epsilon [0, 24)$ & $69.57\% \pm 0.61\%$\\
		\hline
		ImageNet & ResNet-50 & 25.5M & \parbox{3.0cm}{Nesterov SGD \\Weight decay: 0.0001\\batch size: 1024} 
		& $\alpha =\left\{\begin{matrix}
                            0.4\cdot \frac{t}{8} & t\epsilon [0, 8)\\ 
                            0.4& t\epsilon [8, 30)\\ 
                            0.04& t\epsilon [30, 60)\\ 
                            0.004& t\epsilon [60, 80)\\ 
                            0.0004& t\epsilon [80, 90)\\  
                        \end{matrix}\right.$ & $75.54\% \pm 0.01\%$\\
		\bottomrule
	\end{tabular}
\label{tab:dnns}
\end{table*}

\smallskip
\boldhdr{Evaluation Metrics} We evaluate a pruned network on the following three criteria:
\begin{itemize}
    \item \emph{Accuracy:} We use Top-1 accuracy to measure the accuracy of original and pruned models.
    \item \emph{Memory Footprint:} We measure the total memory consumed by the model weights. For pruned models, we consider only the unmasked weights.
    \item \emph{Compute Flops:} We measure the total number of floating point operations (flops) required to perform model inference. For pruned models, we consider only the computation of the unmasked weights.
\end{itemize}

\noindent While comparing accuracy-memory footprint or accuracy-compute flops tradeoff, we also put emphasis on no accuracy loss and 1\% accuracy loss. We observe that higher accuracy losses tend to be unsuitable for industrial applications (like object detection in autonomous driving).

\smallskip
\boldhdr{Baseline} Our baseline is iterative L1-norm weight-based pruning technique~\cite{li2016pruning, renda2020comparing} applied iteratively with rewinding. We refer to it as Iterative L1-norm based pruning (ILP).

\subsection{Accuracy and Memory Footprint Tradeoff}

\begin{figure*}[t]
	\centering
	\subfigure[LeNet-300-100 (MNIST)]{
		\begin{minipage}[t]{0.33\linewidth}
			\centering
			\includegraphics[width=1.8in]{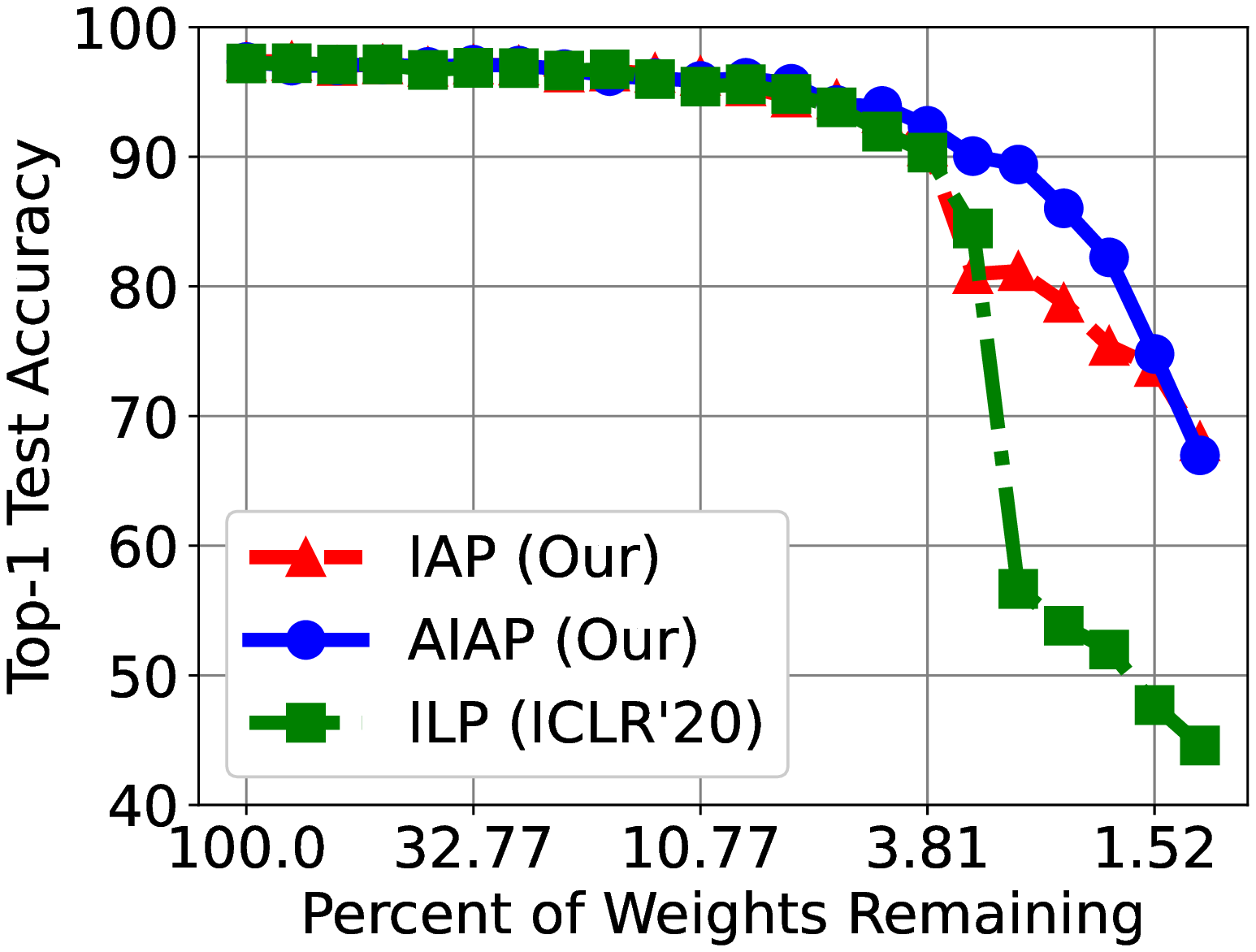}
		\end{minipage}%
	}%
	\subfigure[LeNet-5 (CIFAR-10)]{
		\begin{minipage}[t]{0.33\linewidth}
			\centering
			\includegraphics[width=1.8in]{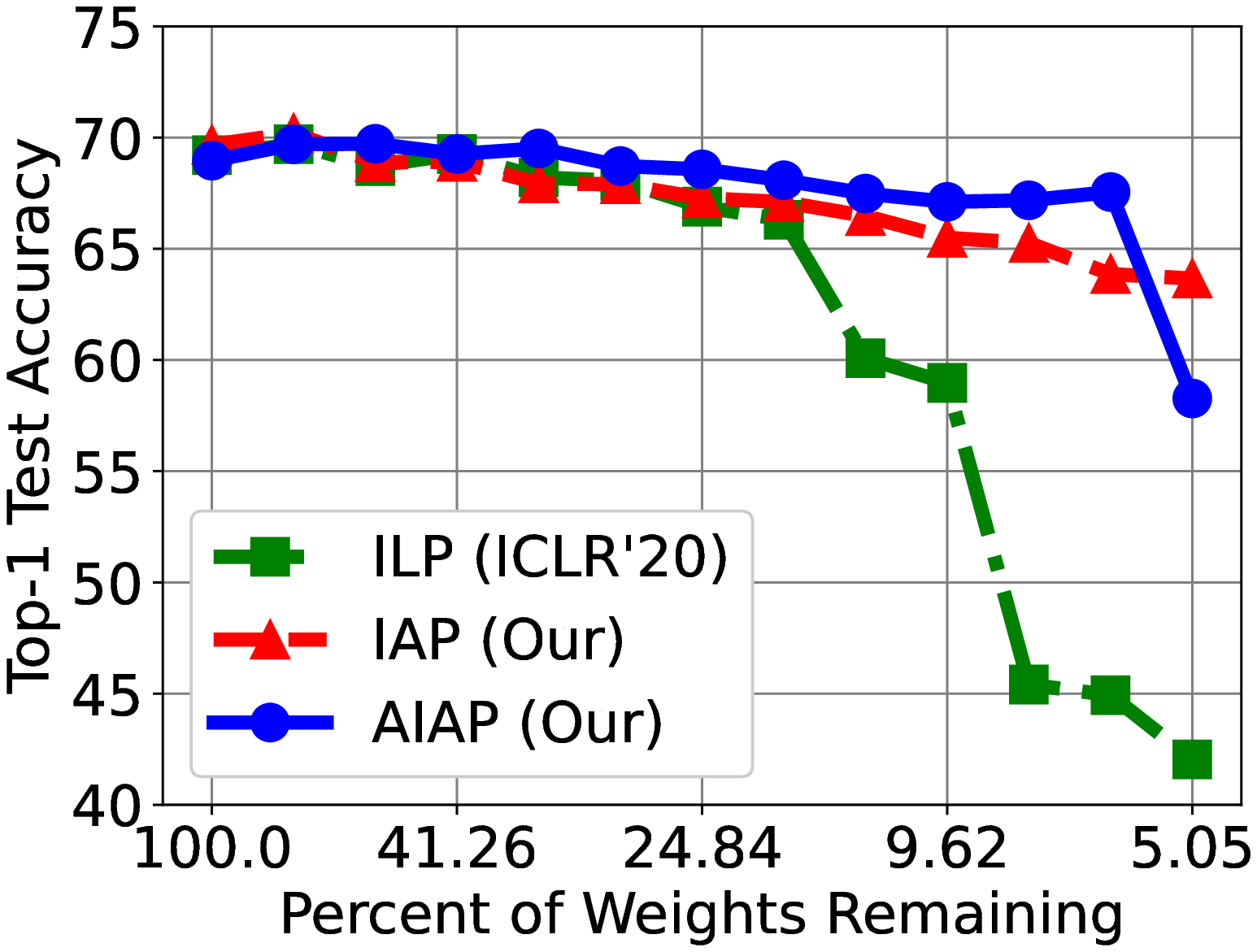}
		\end{minipage}%
	}%
	\subfigure[ResNet-50 (ImageNet)]{
		\begin{minipage}[t]{0.33\linewidth}
			\centering
			\includegraphics[width=1.8in]{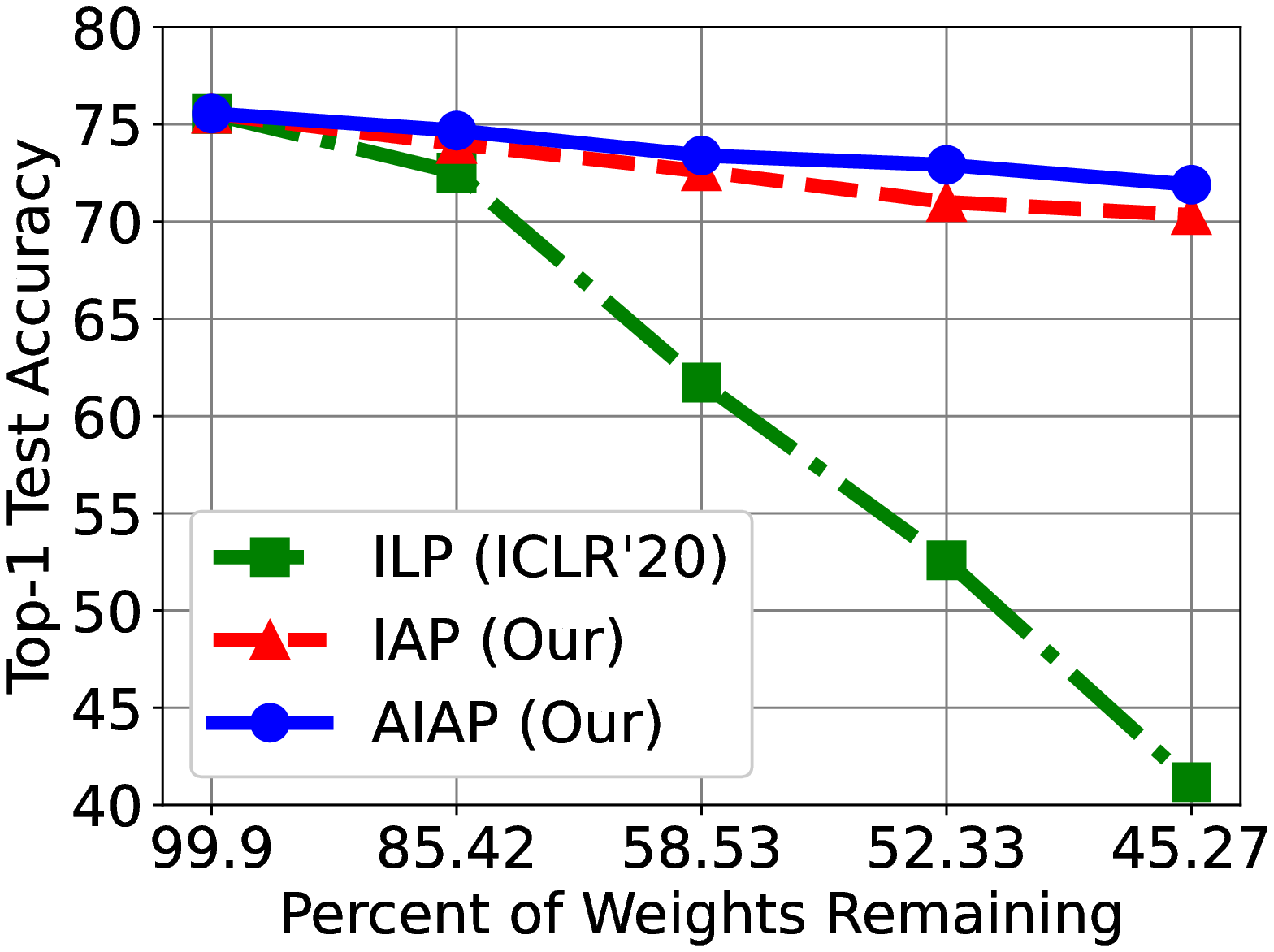}
		\end{minipage}
	}%
	\centering
	\vspace{-6pt}
	\caption{Top-1 test accuracy as the model is pruned iteratively with many pruning rounds. Each pruning round (x-axis) results in a different percentage of remaining number of parameters.}\vspace{-12pt}
	\label{fig: accuracy_vs_parameters}
\end{figure*}

In this subsection, we present the tradeoff between accuracy and memory footprint. We run experiments to seek answers to the first two questions we listed in the Introduction section, 1) Can we find lottery tickets? and 2) What is the accuracy loss with different types of weight- and activation-based pruning techniques?

Referring to Figure~\ref{fig:pruning_flow}, we structurally prune using IAP, AIAP and our baseline ILP. For IAP and AIAP, we use one batch of images (60 for LeNet models and 128
for ResNet-50) on one GPU to calculate activation mean values. For IAP and ILP, in each pruning round, on LeNet models, we prune 20\% for dense layer and 10\% for conv2d layer; On ResNet-50, we prune 5\% for conv2d layer in the first four pruning rounds, and 20\% in later pruning rounds. For AIAP, $\lambda$ is set to 0.01 for LeNet models, and 0.2 for ResNet-50. After pruning, we rewind to roughly 90\% of the total training duration, i.e., rewind to epoch 5 for LeNet-300-100, 22 for LeNet-5 and 80 for ResNet-50 (total training epochs are 6, 24 and 90, respectively). We choose weight rewinding for LeNet models and learning rate rewinding for ResNet-50. For LeNet models, weight rewinding is better than learning rate rewinding (e.g., within 1\% accuracy loss, the pruned model has 12.9\% parameters remaining with weight rewinding and 23.8\% parameters remaining with learning rate rewinding); but their results are similar for ResNet-50. After rewinding, we retrain the models for remaining 10\% of total training epochs. We repeat the cycle for many pruning rounds, and measure the Top-1 test accuracy and the remaining number of unpruned weights at the end of each pruning round.

The results in Figure~\ref{fig: accuracy_vs_parameters} show that IAP and AIAP outperform ILP significantly in all scenarios. On shallow networks, like LeNet-300-100 and LeNet-5, when the number of remaining parameters is large, there is little difference between the pruning methods, and they all match the accuracy of the original network. As the model gets smaller, IAP and AIAP deliver a higher compression at the same accuracy compared to ILP. On LeNet-300-100 with MNIST, with 1.23\% of parameters remaining, the Top-1 accuracy of ILP, IAP, and AIAP is 44.59\%, 68.05\%, and 66.97\%, respectively; On LeNet-5 with CIFAR-10, with 5.05\% remaining parameters, the Top-1 accuracy of ILP, IAP, and AIAP is 42.04\%, 63.66\%, and 58.27\%, respectively.

On deeper models like ResNet-50 with ImageNet, we observe major differences in resulting accuracy of pruned models in early pruning rounds where the compression ratio is not high. IAP and AIAP show a graceful accuracy degradation as the model is pruned iteratively. In comparison, ILP shows steep accuracy losses when the remaining number of weights is below 85.42\%. For example, IAP and AIAP achieve a Top-1 accuracy of 70.33\% and 71.9\%, respectively, with 45.37\% of the parameters remaining, whereas ILP's accuracy drops severely to just 41.17\%.

\begin{table*}[t]
\centering
\small
\caption{Largest achievable compression ratio with 0\% and 1\% accuracy drop}\vspace{-6pt}
\label{tab:compression_ratio}
    \begin{tabular}{@{}lllllll@{}}
    \toprule
                          & \multicolumn{3}{l}{No accuracy drop}  & \multicolumn{3}{l}{1\% accuracy drop} \\ \midrule
    Model                 & IAP (Our) & AIAP (Our) & ILP (ICLR’20) & IAP (Our) & AIAP (Our) & ILP (ICLR’20) \\
    LeNet-300-100 (MNIST) & \textbf{3.80$\times$}  & \textbf{3.80$\times$}  & 1.95$\times$  & 8.94$\times$  & \textbf{9.94$\times$}& 7.42$\times$   \\
    LeNet-5 (CIFAR-10)    & 2.55$\times$  & \textbf{3.95$\times$} & 2.42$\times$  & 7.75$\times$  & \textbf{15.88$\times$} & 4.77$\times$  \\
    ResNet-50 (ImageNet)  & 1.11$\times$ & \textbf{1.17$\times$} & 1.09$\times$          & 1.25$\times$  & \textbf{1.71$\times$}  & 1.13$\times$   
    \\ \bottomrule
    \end{tabular}
\end{table*}

Further, we present the compression ratio for the same experiment at 0\% (no accuracy drop) and 1\% accuracy drop in Table~\ref{tab:compression_ratio}. IAP and AIAP achieve significantly higher compression ratio consistently. For example, with no accuracy loss, on LeNet-5 with CIFAR-10, AIAP can achieve a 3.95$\times$ compression ratio, which is higher than ILP's 2.42$\times$ compression ratio, and so are the results within 1\% accuracy loss (15.88$\times$ vs 4.77$\times$).

With these results, we can answer our research questions:

\begin{enumerate}
    \item \emph{Lottery tickets:} Yes, Table~\ref{tab:compression_ratio} shows that we can get smaller models with no accuracy loss, i.e, lottery tickets, but the compression ratio is small. With an acceptable drop of 1\%, the compression ratio improves significantly, e.g., 1.17$\times$ to 1.71$\times$ for ResNet-50, using AIAP.
    \item \emph{Accuracy loss:} We show that IAP and AIAP produce new state-of-the-art results showing graceful accuracy degradation while iteratively pruning the models.
\end{enumerate}

\subsection{Accuracy and Compute Flops Tradeoff}

\begin{figure*}[t]
	\centering
	\subfigure[LeNet-300-100 (MNIST)]{
		\begin{minipage}[t]{0.33\linewidth}
			\centering
			\includegraphics[width=2.1in]{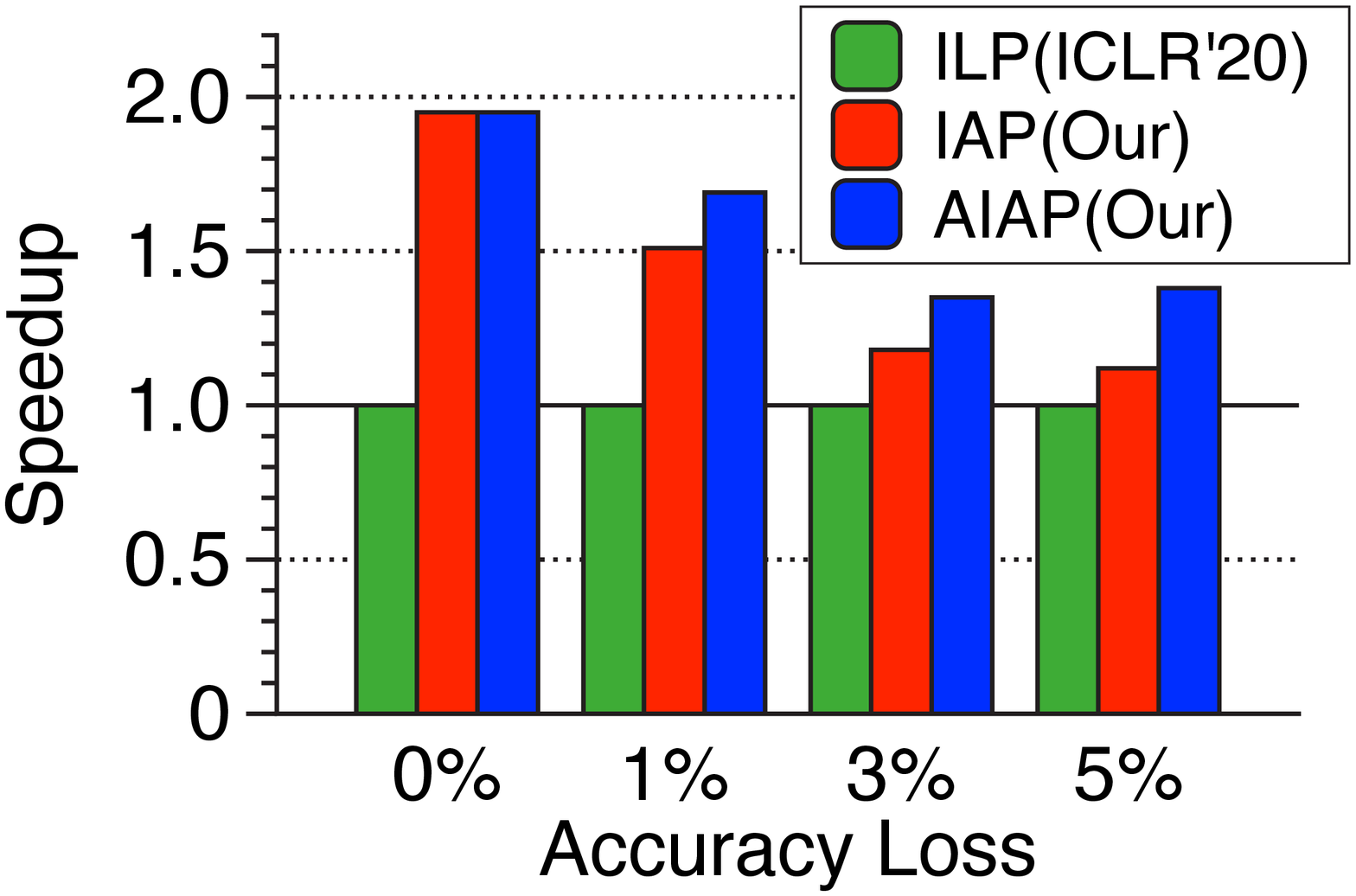}
		\end{minipage}%
	}%
	\subfigure[LeNet-5 (CIFAR-10)]{
		\begin{minipage}[t]{0.33\linewidth}
			\centering
			\includegraphics[width=2.1in]{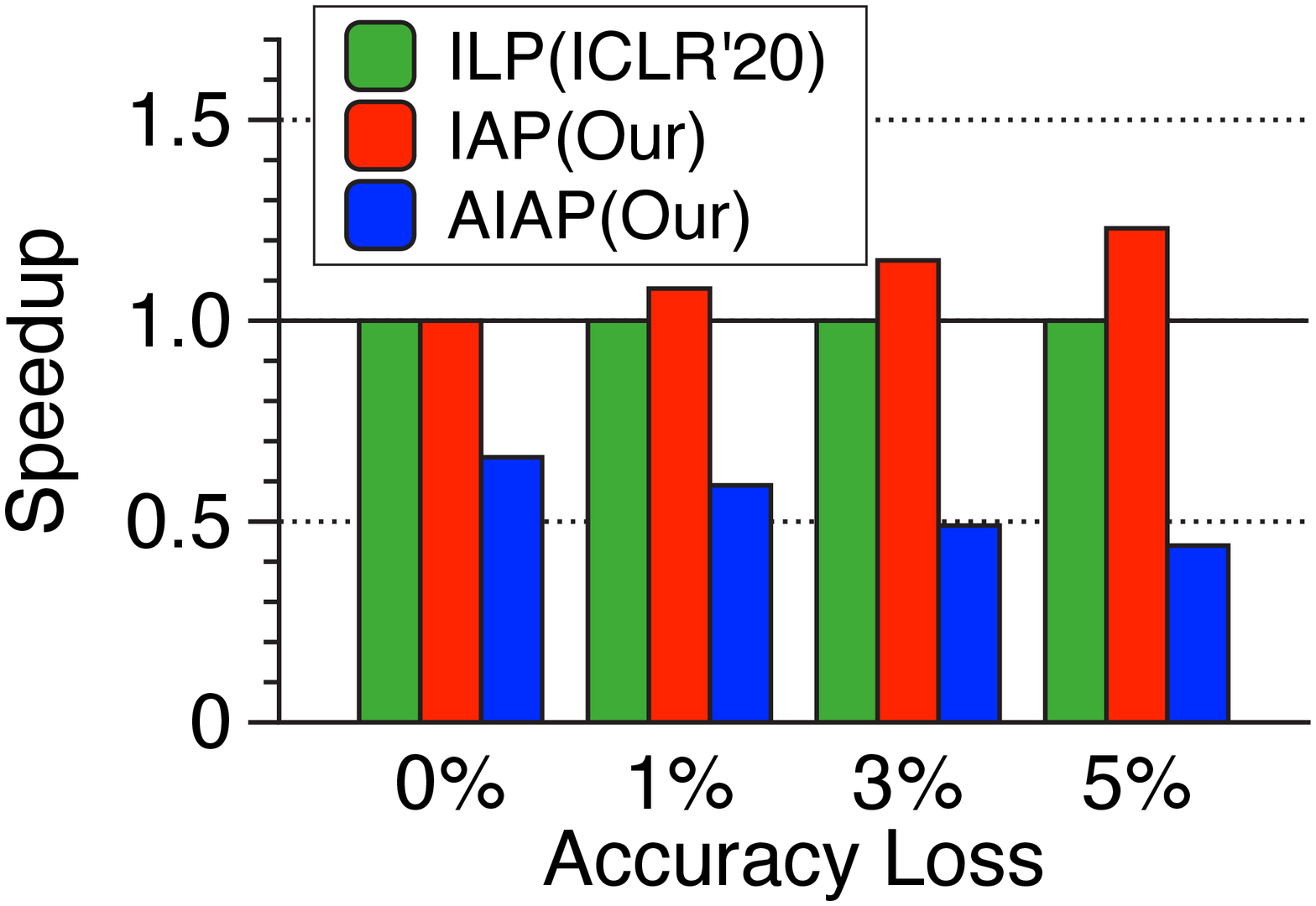}
		\end{minipage}%
	}%
	\subfigure[ResNet-50 (ImageNet)]{
		\begin{minipage}[t]{0.33\linewidth}
			\centering
			\includegraphics[width=2.1in]{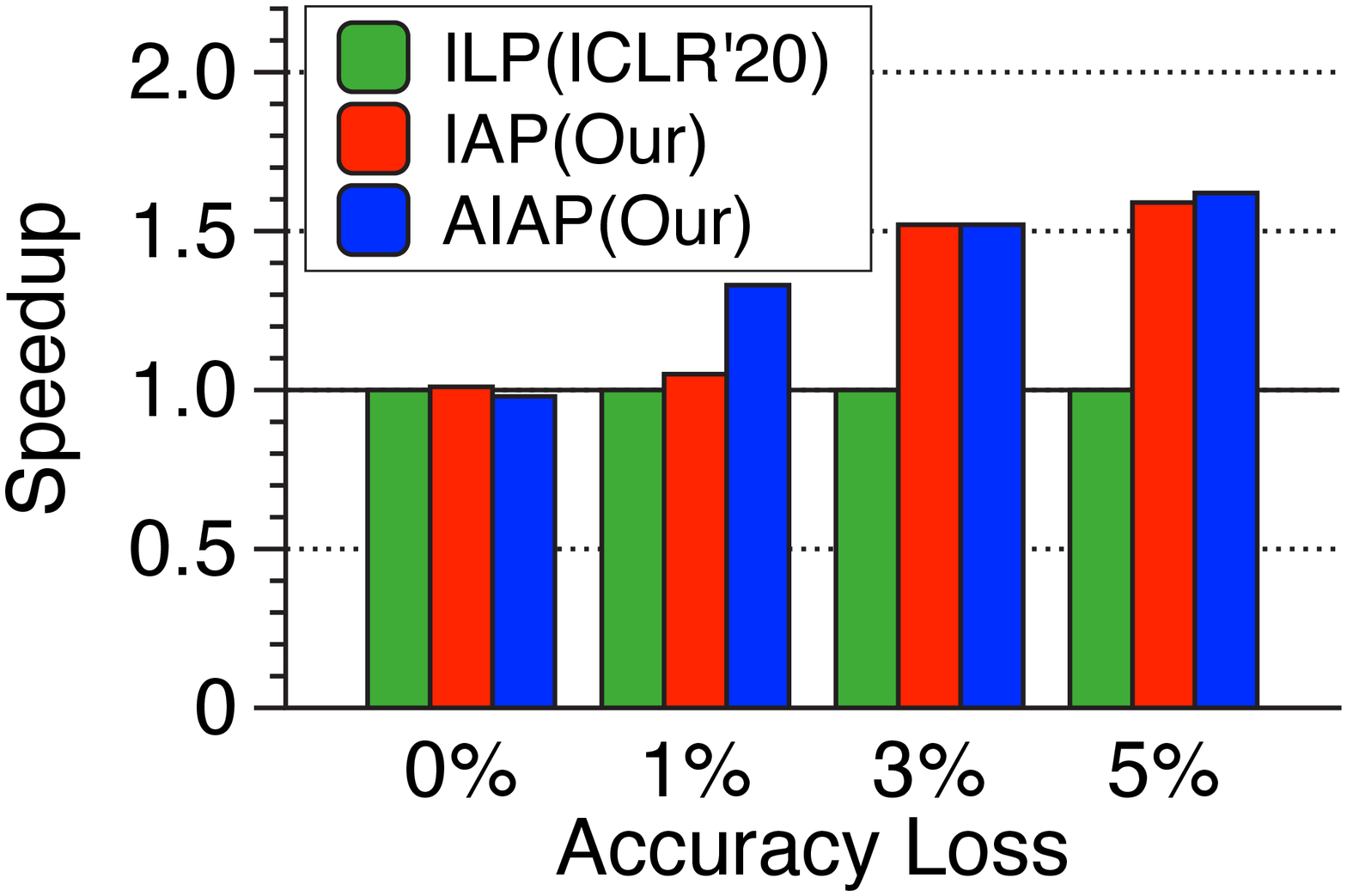}
		\end{minipage}
	}%
	\centering
	\vspace{-6pt}
	\caption{The highest achievable FLOPs speedup for IAP and AIAP normalized to the baseline (ILP), on LeNet-300-100 (MNIST), LeNet-5 (CIFAR-10), and ResNet-50 (ImageNet) with different levels of accuracy loss.}\vspace{-9pt}
	\label{fig: accuracy_vs_flops}
\end{figure*}

Compression ratio alone is not enough to indicate the reduction in inference time. Inference time depends on height and width of the input feature map, in addition to weight tensor sizes. Therefore, we measure the total number of compute flops to estimate the reduction in inference time. Unlike unstructured pruning, compute flops is an acceptable indicator of inference latency for structured pruning because the pruned model has regular patterns and reduces compute flops in a hardware-friendly manner.

In this experiment, we measure the compute flops for the pruned models in our previous experiment across four Top-1 test accuracy drops: 0\%, 1\%, 3\% and 5\%. Figure~\ref{fig: accuracy_vs_flops} shows the total flops for IAP and AIAP normalized to our baseline ILP. We observe that IAP achieves better results in all cases. For example, with no accuracy loss on LeNet-300-100, IAP produces a model that is 1.95$\times$ faster than ILP.

AIAP, however, is not always faster than ILP. It is favorable for LeNet-300-100 and ResNet-50 (0.66$\times$ speedup, with no accuracy loss), but it is worse for LenNet-5 (1.5$\times$ slown down). This is in contrast to the results in Table~\ref{tab:compression_ratio}, where AIAP always achieves higher memory footprint reduction compared to ILP. Upon further investigation, we find that this discrepancy is because AIAP tends to prune more neurons from fully connected layers than conv2d layers. In LeNet-5, conv2d layers are more expensive in terms of compute flops. Therefore for the same compression ratio, ILP (and also IAP) achieve better compute flop reduction.

AIAP achieves a slightly higher speedup than IAP on ResNet-50. This is consistent with our findings in Table~\ref{tab:compression_ratio}, where AIAP achieves higher compression than IAP for the same accuracy loss, e.,g., with 1\% accuracy loss, AIAP achieves higher compression (1.71$\times$ vs 1.25$\times$) and better compute flop reduction (1.33$\times$ vs 1.05$\times$).

These experiments answer our third research question of accuracy and performance tradeoff. We observe that our iterative activation-based pruning techniques are effective in generating pruned models that are not just smaller and accurate, but also \emph{readily} faster.

\subsection{Which Epoch to Rewind?}
In order to understand how rewinding impacts accuracy of the pruned models, we analyze \emph{stability to pruning} as proposed by~\cite{frankle2019stabilizing}, which is defined as the L2 distance between the masked weights of the pruned network and the original network at the end of training.

\begin{figure}[t]
	\centering
	\subfigure[Top-1 Test Accuracy]{
		\begin{minipage}[t]{0.45\linewidth}
			\centering
			\includegraphics[width=1.5in]{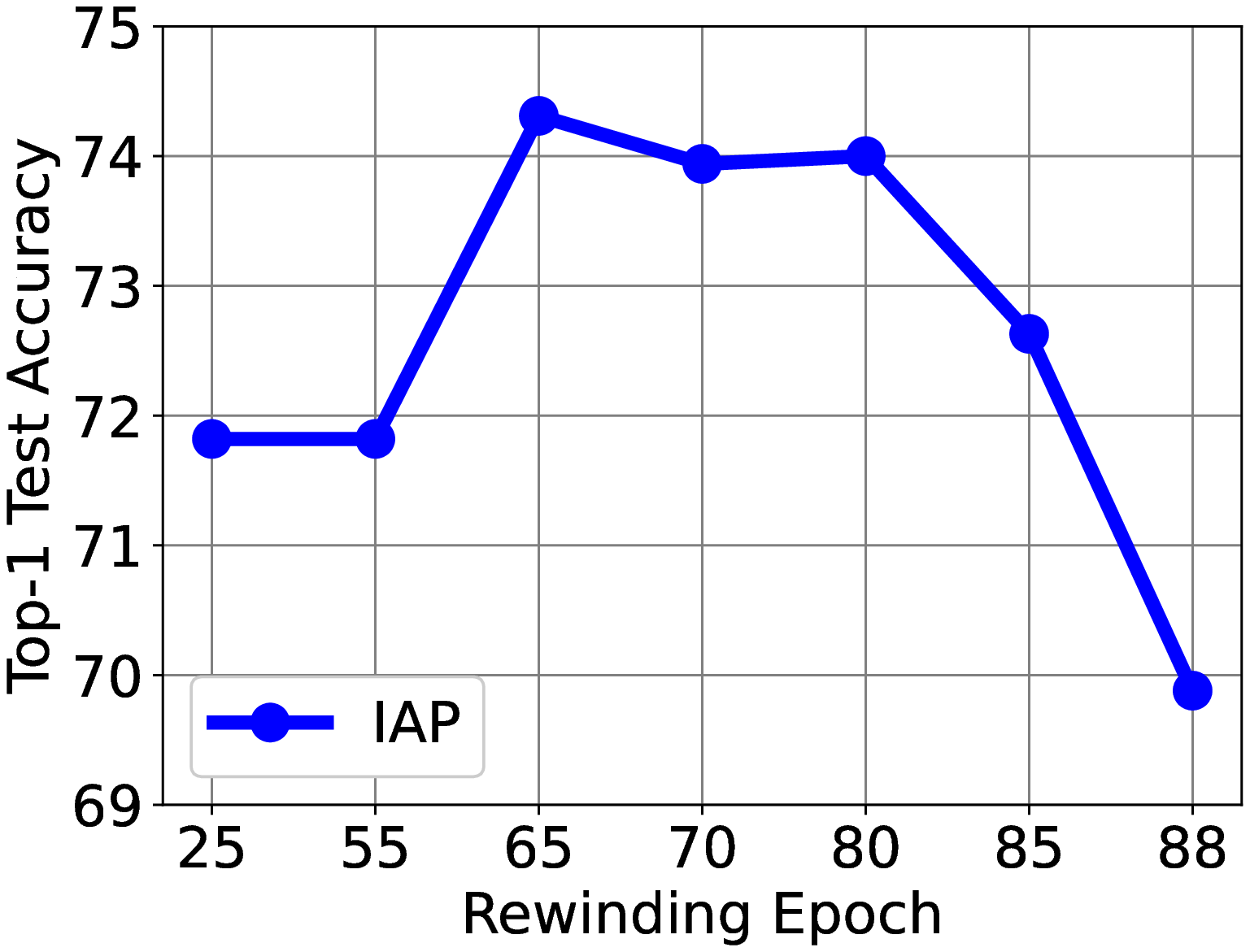}
		\end{minipage}%
	}%
	\subfigure[Stability to Pruning]{
		\begin{minipage}[t]{0.45\linewidth}
			\centering
			\includegraphics[width=1.5in]{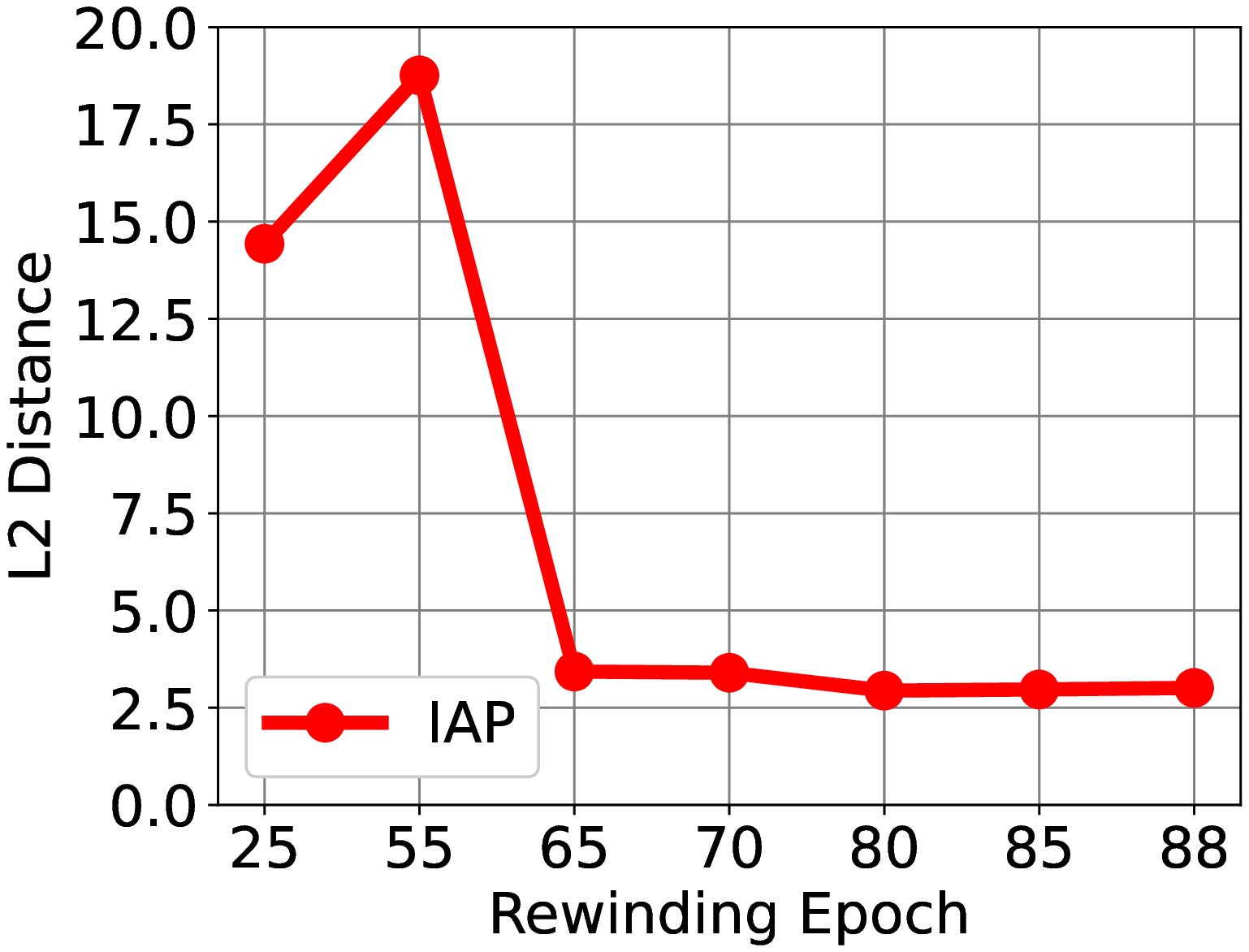}
		\end{minipage}%
	}%
	\centering
	\vspace{-6pt}
	\caption{The effect of the rewinding epoch (x-axis) on Top-1 test accuracy (left) and pruning stability (right) for pruned ResNet-50, when 83.74\% of parameters are remaining.}\vspace{-9pt}
	\label{fig: stability}
\end{figure}

Figure~\ref{fig: stability}(a) shows the Top-1 test accuracy of the pruned ResNet-50 with 83.74\% remaning parameters when learning rate is rewound to different epochs, and Figure~\ref{fig: stability}(b) shows the stability values at the corresponding rewinding epochs. We observe that there is a region, 65 to 80 epochs, where the resulting accuracy is high. Below this range, the pruned model accuracy is sub-optimal. We find that L2 distance closely follows this pattern, showing high distance for early training epochs and small distance for later training epochs. We also observe that when rewinding epoch is very close to the end of training time (85 and 88 epochs), the resulting accuracy is bad.

We therefore validate the observations proposed by~\cite{frankle2019stabilizing} that for deep networks, rewinding to very early stages is sub-optimal as the network has not learned considerably by then. On the other hand, rewinding to very late training stages is also sub-optimal because there is not enough time to retrain. Our findings show that rewinding to 75\%-90\% of training time leads to good accuracy.

\section{Conclusion}\label{conclusion}
In this paper, we investigate iterative structured pruning with rewinding. Structural pruning removes entire filters from  weight tensors, resulting in pruned models that can be easily accelerated using off-the-shelf libraries. We introduce two activation-based iteration structured pruning techniques---Iterative Activation-based Pruning (IAP) and Adaptive Iterative Activation-based Pruning (AIAP) to aggressively prune the models. We observe that IAP and AIAP can find winning lottery tickets, i.e., models that are smaller and more accurate than the original model. Our methods also substantially outperform weight-based Iterative L1-norm Pruning (ILP) on a variety of architectures, e.g., LeNet-300-100 (MNIST), LeNet-5 (CIFAR-10), and ResNet-50 (ImageNet). For the same memory footprint, IAP and AIAP achieve much higher accuracy (e.g., 30\% higher Top-1 accuracy for pruned ResNet-50 models with 45.27\% remaining parameters); for the same accuracy loss, they achieve much higher compression and faster inference speed (e.g., 1.95$\times$ higher compression and 1.95$\times$ speedup for pruned LeNet-300-100 models with no accuracy loss).



    \bibliography{reference}

\begin{thebibliography}{31}
\providecommand{\natexlab}[1]{#1}
\providecommand{\url}[1]{\texttt{#1}}
\providecommand{\urlprefix}{URL }
\expandafter\ifx\csname urlstyle\endcsname\relax
  \providecommand{\doi}[1]{doi:\discretionary{}{}{}#1}\else
  \providecommand{\doi}{doi:\discretionary{}{}{}\begingroup
  \urlstyle{rm}\Url}\fi

\bibitem[{Chu, Funderlic, and Plemmons(2003)}]{chu2003structured}
Chu, M.~T.; Funderlic, R.~E.; and Plemmons, R.~J. 2003.
\newblock Structured low rank approximation.
\newblock \emph{Linear algebra and its applications} 366: 157--172.

\bibitem[{Clarkson and Woodruff(2017)}]{clarkson2017low}
Clarkson, K.~L.; and Woodruff, D.~P. 2017.
\newblock Low-rank approximation and regression in input sparsity time.
\newblock \emph{Journal of the ACM (JACM)} 63(6): 1--45.

\bibitem[{Deng et~al.(2009)Deng, Dong, Socher, Li, Li, and
  Fei-Fei}]{deng2009imagenet}
Deng, J.; Dong, W.; Socher, R.; Li, L.-J.; Li, K.; and Fei-Fei, L. 2009.
\newblock Imagenet: A large-scale hierarchical image database.
\newblock In \emph{2009 IEEE conference on computer vision and pattern
  recognition}, 248--255. Ieee.

\bibitem[{Deng(2012)}]{deng2012mnist}
Deng, L. 2012.
\newblock The mnist database of handwritten digit images for machine learning
  research [best of the web].
\newblock \emph{IEEE Signal Processing Magazine} 29(6): 141--142.

\bibitem[{Elsen et~al.(2019)Elsen, Dukhan, Gale, and Simonyan}]{sparse_conv}
Elsen, E.; Dukhan, M.; Gale, T.; and Simonyan, K. 2019.
\newblock Fast Sparse ConvNets.

\bibitem[{Frankle and Carbin(2018)}]{frankle2018lottery}
Frankle, J.; and Carbin, M. 2018.
\newblock The lottery ticket hypothesis: Finding sparse, trainable neural
  networks.
\newblock \emph{arXiv preprint arXiv:1803.03635} .

\bibitem[{Frankle et~al.(2019)Frankle, Dziugaite, Roy, and
  Carbin}]{frankle2019stabilizing}
Frankle, J.; Dziugaite, G.~K.; Roy, D.~M.; and Carbin, M. 2019.
\newblock Stabilizing the lottery ticket hypothesis.
\newblock \emph{arXiv preprint arXiv:1903.01611} .

\bibitem[{Han, Mao, and Dally(2015)}]{han2015deep}
Han, S.; Mao, H.; and Dally, W.~J. 2015.
\newblock Deep compression: Compressing deep neural networks with pruning,
  trained quantization and huffman coding.
\newblock \emph{arXiv preprint arXiv:1510.00149} .

\bibitem[{Han et~al.(2015)Han, Pool, Tran, and Dally}]{han2015learning}
Han, S.; Pool, J.; Tran, J.; and Dally, W. 2015.
\newblock Learning both weights and connections for efficient neural network.
\newblock In \emph{Advances in neural information processing systems},
  1135--1143.

\bibitem[{He et~al.(2016)He, Zhang, Ren, and Sun}]{he2016deep}
He, K.; Zhang, X.; Ren, S.; and Sun, J. 2016.
\newblock Deep residual learning for image recognition.
\newblock In \emph{Proceedings of the IEEE conference on computer vision and
  pattern recognition}, 770--778.

\bibitem[{He et~al.(2018)He, Lin, Liu, Wang, Li, and Han}]{he2018amc}
He, Y.; Lin, J.; Liu, Z.; Wang, H.; Li, L.-J.; and Han, S. 2018.
\newblock Amc: Automl for model compression and acceleration on mobile devices.
\newblock In \emph{Proceedings of the European Conference on Computer Vision
  (ECCV)}, 784--800.

\bibitem[{{Hill} et~al.(2017){Hill}, {Jain}, {Hill}, {Zamirai}, {Hsu},
  {Laurenzano}, {Mahlke}, {Tang}, and {Mars}}]{deftnn}
{Hill}, P.; {Jain}, A.; {Hill}, M.; {Zamirai}, B.; {Hsu}, C.; {Laurenzano},
  M.~A.; {Mahlke}, S.; {Tang}, L.; and {Mars}, J. 2017.
\newblock DeftNN: Addressing Bottlenecks for DNN Execution on GPUs via Synapse
  Vector Elimination and Near-compute Data Fission.
\newblock In \emph{2017 50th Annual IEEE/ACM International Symposium on
  Microarchitecture (MICRO)}, 786--799.

\bibitem[{Howard et~al.(2017)Howard, Zhu, Chen, Kalenichenko, Wang, Weyand,
  Andreetto, and Adam}]{howard2017mobilenets}
Howard, A.~G.; Zhu, M.; Chen, B.; Kalenichenko, D.; Wang, W.; Weyand, T.;
  Andreetto, M.; and Adam, H. 2017.
\newblock Mobilenets: Efficient convolutional neural networks for mobile vision
  applications.
\newblock \emph{arXiv preprint arXiv:1704.04861} .

\bibitem[{Kadetotad et~al.(2016)Kadetotad, Arunachalam, Chakrabarti, and
  Seo}]{kadetotad2016efficient}
Kadetotad, D.; Arunachalam, S.; Chakrabarti, C.; and Seo, J.-s. 2016.
\newblock Efficient memory compression in deep neural networks using
  coarse-grain sparsification for speech applications.
\newblock In \emph{Proceedings of the 35th International Conference on
  Computer-Aided Design}, 78. ACM.

\bibitem[{Krizhevsky, Hinton et~al.(2009)}]{krizhevsky2009learning}
Krizhevsky, A.; Hinton, G.; et~al. 2009.
\newblock Learning multiple layers of features from tiny images.
\newblock Technical report, Citeseer.

\bibitem[{Lan, Studer, and Baraniuk(2014)}]{lan2014matrix}
Lan, A.~S.; Studer, C.; and Baraniuk, R.~G. 2014.
\newblock Matrix recovery from quantized and corrupted measurements.
\newblock In \emph{2014 IEEE International Conference on Acoustics, Speech and
  Signal Processing (ICASSP)}, 4973--4977. IEEE.

\bibitem[{LeCun, Denker, and Solla(1990)}]{lecun1990optimal}
LeCun, Y.; Denker, J.~S.; and Solla, S.~A. 1990.
\newblock Optimal brain damage.
\newblock In \emph{Advances in neural information processing systems},
  598--605.

\bibitem[{Lee, Ajanthan, and Torr(2018)}]{lee2018snip}
Lee, N.; Ajanthan, T.; and Torr, P.~H. 2018.
\newblock Snip: Single-shot network pruning based on connection sensitivity.
\newblock \emph{arXiv preprint arXiv:1810.02340} .

\bibitem[{Li et~al.(2019)Li, Zeng, Zhou, and Chen}]{li2019edge}
Li, E.; Zeng, L.; Zhou, Z.; and Chen, X. 2019.
\newblock Edge AI: On-demand accelerating deep neural network inference via
  edge computing.
\newblock \emph{IEEE Transactions on Wireless Communications} 19(1): 447--457.

\bibitem[{Li, Zhou, and Chen(2018)}]{li2018edge}
Li, E.; Zhou, Z.; and Chen, X. 2018.
\newblock Edge intelligence: On-demand deep learning model co-inference with
  device-edge synergy.
\newblock In \emph{Proceedings of the 2018 Workshop on Mobile Edge
  Communications}, 31--36.

\bibitem[{Li et~al.(2016)Li, Kadav, Durdanovic, Samet, and
  Graf}]{li2016pruning}
Li, H.; Kadav, A.; Durdanovic, I.; Samet, H.; and Graf, H.~P. 2016.
\newblock Pruning filters for efficient convnets.
\newblock \emph{ICLR} .

\bibitem[{Molchanov et~al.(2016)Molchanov, Tyree, Karras, Aila, and
  Kautz}]{molchanov2016pruning}
Molchanov, P.; Tyree, S.; Karras, T.; Aila, T.; and Kautz, J. 2016.
\newblock Pruning convolutional neural networks for resource efficient
  inference.
\newblock \emph{arXiv preprint arXiv:1611.06440} .

\bibitem[{Pham et~al.(2018)Pham, Guan, Zoph, Le, and Dean}]{pham2018efficient}
Pham, H.; Guan, M.~Y.; Zoph, B.; Le, Q.~V.; and Dean, J. 2018.
\newblock Efficient neural architecture search via parameter sharing.
\newblock \emph{arXiv preprint arXiv:1802.03268} .

\bibitem[{Polino, Pascanu, and Alistarh(2018)}]{polino2018model}
Polino, A.; Pascanu, R.; and Alistarh, D. 2018.
\newblock Model compression via distillation and quantization.
\newblock \emph{arXiv preprint arXiv:1802.05668} .

\bibitem[{Renda, Frankle, and Carbin(2020)}]{renda2020comparing}
Renda, A.; Frankle, J.; and Carbin, M. 2020.
\newblock Comparing rewinding and fine-tuning in neural network pruning.
\newblock \emph{arXiv preprint arXiv:2003.02389} .

\bibitem[{Srinivas and Babu(2015)}]{srinivas2015data}
Srinivas, S.; and Babu, R.~V. 2015.
\newblock Data-free parameter pruning for deep neural networks.
\newblock \emph{arXiv preprint arXiv:1507.06149} .

\bibitem[{Tan and Le(2019)}]{tan2019efficientnet}
Tan, M.; and Le, Q.~V. 2019.
\newblock Efficientnet: Rethinking model scaling for convolutional neural
  networks.
\newblock \emph{arXiv preprint arXiv:1905.11946} .

\bibitem[{Theis et~al.(2018)Theis, Korshunova, Tejani, and
  Husz{\'a}r}]{theis2018faster}
Theis, L.; Korshunova, I.; Tejani, A.; and Husz{\'a}r, F. 2018.
\newblock Faster gaze prediction with dense networks and fisher pruning.
\newblock \emph{arXiv preprint arXiv:1801.05787} .

\bibitem[{Yim et~al.(2017)Yim, Joo, Bae, and Kim}]{yim2017gift}
Yim, J.; Joo, D.; Bae, J.; and Kim, J. 2017.
\newblock A gift from knowledge distillation: Fast optimization, network
  minimization and transfer learning.
\newblock In \emph{Proceedings of the IEEE Conference on Computer Vision and
  Pattern Recognition}, 4133--4141.

\bibitem[{Yu et~al.(2018)Yu, Yang, Xu, Yang, and Huang}]{yu2018slimmable}
Yu, J.; Yang, L.; Xu, N.; Yang, J.; and Huang, T. 2018.
\newblock Slimmable neural networks.
\newblock \emph{arXiv preprint arXiv:1812.08928} .

\bibitem[{Zoph and Le(2016)}]{zoph2016neural}
Zoph, B.; and Le, Q.~V. 2016.
\newblock Neural architecture search with reinforcement learning.
\newblock \emph{arXiv preprint arXiv:1611.01578} .

\end{thebibliography}

\end{document}